\documentclass{llncs}

\def\techreport{}

\usepackage{graphicx}
\usepackage{mathtools, amsmath, amssymb}
\usepackage{algorithm}
\usepackage{ifthen}
\usepackage{color}
\usepackage{tikz}
\usepackage{multirow}
\usepackage[format=hang]{subfig}
\usepgflibrary{shapes}
\usetikzlibrary{arrows,positioning}
\newcommand{\tool}{{\sc MultiGain}}

\newcommand{\ov}{\overline}
\newcommand{\wh}{\widehat}
\newcommand{\src}{\mathit{Src}}
\title{\tool: A controller synthesis tool for MDPs with multiple mean-payoff objectives}
\author{
Tom\'a\v{s} Br\'azdil\inst{1}\!\and
Krishnendu~Chatterjee\inst{2} \and
Vojt\v{e}ch~Forejt\inst{3} \and
Anton\'in Ku\v{c}era \inst{1}
}

\institute{
Faculty of Informatics, Masaryk University, Brno, Czech Republic
\and
IST Austria
\and
Department of Computer Science, University of Oxford, UK
}

\begin{document}

\maketitle
\begin{abstract}
We present \tool, a tool to synthesize strategies for Markov decision 
processes (MDPs) with multiple mean-payoff objectives.
Our models are described in PRISM, and our tool uses the existing interface 
and simulator of PRISM.
Our tool extends PRISM by adding novel algorithms for multiple mean-payoff 
objectives, and also provides features such as (i)~generating strategies and 
exploring them for simulation, and checking them with respect to other properties; 
and (ii)~generating an approximate Pareto curve for two mean-payoff objectives.
In addition, we present a new practical algorithm for the analysis of MDPs
with multiple mean-payoff objectives under memoryless strategies.
\end{abstract}

\newcommand{\act}[1]{\mathit{Act}(#1)}
\newcommand{\Exp}{\mathbb{E}}
\newcommand{\dist}{\mathit{dist}}
\newcommand{\mpo}{\mathsf{mp}}

\section{Introduction}
\emph{Markov decision processes (MDPs)} are the de facto model 
for analysis of probabilistic systems with non-determinism~\cite{Howard}, with a wide range of applications~\cite{BaierBook}.
In each state of an MDP, a controller chooses one of several actions 
(the nondeterministic choices), and the current state and action 
gives a probability distribution over the successor states.
%MDPs have been widely used in diverse domains~\cite{BaierBook,KNP11}
%ranging from analysis of randomized communication and security protocols 
%to biological systems.
%
One classical objective used to study quantitative properties
of systems is the \emph{limit-average (or mean-payoff)} objective, where a 
reward (or cost) is associated with each transition and the objective 
assigns to every run the average of the rewards over the run.
MDPs with single mean-payoff objectives have been well studied in the 
literature (see, e.g., \cite{Puterman}).
However, in many modeling domains, there is not a single goal to be 
optimized, but multiple, potentially interdependent and conflicting goals.
For example, in designing a computer system, the goal is to maximize 
average performance while minimizing average power consumption.
Similarly, in an inventory management system, the goal is to optimize
several dependent costs for maintaining each kind of product.
The complexity of MDPs with multiple mean-payoff objectives was studied in~\cite{BBC+11}.

In this paper we present \tool, which is, to the best of our knowledge, the first 
tool for synthesis of controller strategies in MDPs with multiple mean-payoff
objectives. 
The MDPs and the mean-payoff objectives are specified in the well-known 
PRISM modelling language.
Our contributions are as follows: 
(1)~we extend PRISM with novel algorithms for multiple mean-payoff objectives 
from~\cite{BBC+11}; (2)~develop on the results of~\cite{BBC+11} to synthesize 
strategies, and explore them for simulation, and check them with respect to 
other properties (as done in PRISM-games~\cite{CFK+13}); and (3)~for the important special 
case of two mean-payoff objectives we provide the feature to
visualize the approximate Pareto curve 
(where the Pareto curve represents the ``trade-off'' curve and consists of solutions
that are not strictly dominated by any other solution).
Finally, we present a new practical approach for analysis of MDPs with multiple
mean-payoff objectives under memoryless strategies: previously an NP bound was 
shown in~\cite{Cha07} by guessing all bottom strongly connected components (BSCCs) 
of the MDP graph for a memoryless strategy and this gave an exponential enumerative 
algorithm; in contrast, we present a linear reduction to solving a boolean combination 
of linear constraints (which is a special class of mixed integer linear programming where 
the integer variables are binary).

\newcommand{\pat}{\pi}
\vspace{-0.5em}
\section{Definitions}
\vspace{-0.5em}

\noindent{\bf MDPs and strategies.} 
An MDP  $G=(S,A,\mathit{Act},\delta)$ consists of
(i)~a \emph{finite} set $S$ of states; (ii)~a \emph{finite} set $A$ of actions, 
(iii)~an action enabledness function $\mathit{Act} : S\rightarrow 2^A\setminus 
\{\emptyset\}$ that assigns to each state $s$ the set $\act{s}$ of actions 
enabled at $s$, and (iv)~a transition function 
$\delta : S\times A\rightarrow \dist(S)$ that given a state $s$ and an action 
$a \in \act{s}$ gives a probability distribution over the successor states 
($\dist(S)$ denotes all probability distributions over $S$).
W.l.o.g. we assume that every action is enabled in exactly one state, 
and we denote this state $\src(a)$. 
Thus, we will assume that $\delta: A \rightarrow \dist(S)$.
{\em Strategies} describe how 
to choose the next action given a finite path (of state and action pairs) in 
the MDP.
A strategy consists of a set of memory elements to remember the history of the 
paths.
The memory elements are updated stochastically in each transition, and the 
next action is chosen probabilistically (among enabled actions) based on the 
current state and current memory~\cite{BBC+11}.
A strategy is
%\emph{finite-memory} if the number of memory elements used is 
%finite, and is
\emph{memoryless} if it
%is independent of the history, but 
depends only on the current state.

\smallskip\noindent{\bf Multiple mean-payoff objectives.} 
A single mean-payoff objective consists of a reward function $r$ that assigns a 
real-valued reward $r(s,a)$ to every state $s$ and action $a$ enabled in $s$, 
and the mean-payoff objective $\mpo(r)$ assigns to every infinite path (or run) the long-run 
average of the rewards of the path, i.e., for a infinite path $\pat=(s_0 a_0 s_1 a_1 \ldots)$ 
we have $\mpo(r)(\pat)= \lim\inf_{n \to \infty} \frac{1}{n} \cdot \sum_{i=0}^{n-1} r(s_i,a_i)$.
In multiple mean-payoff objectives, there are $k$ reward functions $r_1, r_2,\ldots, r_k$,
and each reward function $r_i$ defines the respective mean-payoff objective $\mpo(r_i)$.
Given a strategy $\sigma$, we denote by $\Exp_{s}^{\sigma}[\cdot]$ the 
expectation measure of the strategy given a starting state $s$.
Thus for a mean-payoff objective $\mpo(r)$, the expected mean-payoff is
$\Exp_{s}^{\sigma}[\mpo(r)]$.

\smallskip\noindent{\bf Synthesis questions.} 
The relevant questions in analysis of MDPs with multiple objectives are as follows:
(1)~\emph{(Existence).} Given an MDP with $k$ reward functions, starting state $s_0$, 
and a vector $\vec{v}=(v_1,v_2,\ldots,v_k)$ of $k$ real-values, the existence question asks 
whether there exists a strategy $\sigma$ such that for all $1 \leq i \leq k$ we have 
$\Exp_{s_0}^{\sigma}[\mpo(r_i)] \geq v_i$.
(2)~\emph{(Synthesis).} If the answer to the existence question is yes, the synthesis
question asks for a witness strategy to satisfy the existence question.
An optimization question related to multiple objectives is the computation of the 
Pareto-curve (or the trade-off curve), where the Pareto curve consists of 
vectors $\vec{v}$ such that the answer to the existence question is yes, and for all
vectors $\vec{v}'$ that strictly dominate $\vec{v}$ (i.e., $\vec{v}'$ is at least 
$\vec{v}$ in all dimensions and strictly greater in at least one dimension) 
the answer to the existence question is no.

\section{Algorithms and Implementation}
We first recall the existing results for MDPs with multiple 
mean-payoff objectives~\cite{BBC+11},
and then describe our implementation and extensions.
Before presenting the existing results, we first recall the notion 
of maximal end-components in MDPs.

\smallskip\noindent{\bf Maximal end-components.}
A pair $(T,B)$ with $\emptyset\neq T\subseteq S$ and $B\subseteq \bigcup_{t\in T}\act{t}$
is an \emph{end component} of $G$ if (1) for all $a\in B$, whenever $\delta(a)(s')>0$ then $s'\in T$;
and (2) for all $s,t\in T$ there is a finite path from $s$ to $t$ such that  
all states and actions that appear in the path belong to $T$ and $B$, respectively.
An end component $(T,B)$ is a \emph{maximal end component (MEC)}
if it is maximal wrt.\ pointwise subset ordering. 
An MDP is \emph{unichain} if for all $B \subseteq A$ satisfying 
$B \cap \act{s} \neq \emptyset$ for any $s \in S$ we have that $(S,B)$ is a MEC.
Given an MDP, we denote  $S_{\textit{MEC}}$ the set of states $s$ that are contained within a MEC.

\smallskip\noindent{\bf Result from~\cite{BBC+11}.} The results of~\cite{BBC+11} showed that 
(i)~the existence question can be answered in polynomial time, by reduction to linear programming;
(ii)~if there exists a strategy for the existence problem, then there exists a witness strategy 
with only two-memory states.
It also established that if the MDP is unichain, then memoryless strategies are 
sufficient. 
The polynomial-time algorithm is as follows:
it was shown in~\cite{BBC+11} that the answer to the existence problem is yes iff there exists a 
non-negative solution to the system of linear inequalities given in Fig.~\ref{system-L}.

\begin{figure}[t]\small
\begin{align}
\mathbf{1}_{s_0}(s) + \textstyle\sum\nolimits_{a\in A} y_{a}\cdot \delta(a)(s) & = 
 \textstyle\sum\nolimits_{a\in \act{s}} y_{a} + y_s &\text{for all  $s\in S$}
\label{eq:ya}\\
 % && 
% \llap{\text{
% for all  $s\in S$;
% here
% $init(s)=\begin{cases}
% 1 & s=s_0\\
% 0 & s\neq s_0
% \end{cases}$
% }}
% \notag\\
\textstyle\sum\nolimits_{s\in S_{\textit{MEC}}}y_{s} & =  1 &
\label{eq:ys1}\\
 \textstyle\sum\nolimits_{s\in C} y_{s} & =  \textstyle\sum\nolimits_{a\in A\cap C} x_{a} &
   \text{for all MECs $C$ of $G$}
\label{eq:yC}\\
%  && 
% \llap{\text{
% for all MEC $C$ of $G$
% }
% }
% \notag
% \\
\textstyle\sum\nolimits_{a\in A} x_{a}\cdot \delta(a)(s) & = 
\textstyle\sum\nolimits_{a\in \act{s}} x_{a} & \text{for all  $s\in S$}
\label{eq:xa}\\
% &&
% \llap{
% \text{ for all $s\in S$}}
% \notag
% \\
\textstyle\sum\nolimits_{a\in A} x_{a}\cdot\vec{r}_i(a) & \ge  \vec{v}_i &
\text{ for all $1\le i \le k$}
\label{eq:rew}
\end{align}
\caption{System $L$ of linear inequalities (here
$\mathbf{1}_{s_0}(s)$ is $1$ 
if $s {=} s_0$, and $0$ otherwise).}
\label{system-L}
\vspace{-1em}
\end{figure}

\smallskip\noindent{\bf Syntax and semantics.} Our tool accepts PRISM MDP models as input, see
\cite{prism-web} for details. The multi-objective properties are expressed as
\texttt{multi(}\textit{list}\texttt{)}
or
\texttt{mlessmulti(}\textit{list}\texttt{)}
where \textit{list} is a comma separated list of mean-payoff reward properties, which can
be {\em boolean}, e.g. \texttt{R\{'r1'\}>=0.5 [S]}, and
in the case of \texttt{multi} also {\em numerical}, e.g. \texttt{R\{'r2'\}min=? [S]}.
In the reward properties, \texttt{S} stands for steady-state,
following PRISM's terminology.

If all properties in the list are boolean, the multi-objective
property \texttt{multi(}\textit{list}\texttt{)} is also boolean and is true iff there is a strategy
under which all given reward properties in the list are simultaneously satisfied. If there is a single
numerical query, the multi-objective query intuitively asks for the maximal achievable reward of the
numerical reward query, subject to the restriction given by the boolean queries. We also allow
two numerical queries; in such case \tool{} generates a Pareto curve.
The semantics of \texttt{mlessmulti} follows the same pattern, the only difference being that only
memoryless (randomised) strategies are being considered. The reason we don't allow
numerical reward properties in \texttt{mlessmulti} is that the supremum among all memoryless
strategies might not be realised.

\smallskip\noindent{\bf Implementation of existence question.}
We have implemented the algorithm of~\cite{BBC+11}. 
Our implementation takes as input an MDP 
with multiple mean-payoff objectives and a value vector $\vec{v}$,
and computes the linear inequalities of Fig.~\ref{system-L} or a {\em mixed integer linear programming} (MILP) extension
in case of memoryless strategies.
The system of linear inequalities is solved with LPsolve~\cite{lpsolve} or Gurobi~\cite{gurobi}.

\smallskip\noindent{\bf Implementation of the synthesis question.}
We now describe how to obtain witness strategies. Assume that the linear program
from Fig.~\ref{system-L} has a solution, where a solution to a variable $z$ is denoted by $\ov{z}$.
We construct a new linear program, comprising Eq.~\ref{eq:ya}
together with the equations
$
 y_s = \textstyle\sum\nolimits_{a\in \act{s}} \ov{x}_a
$
for all $s\in S_{\textit{MEC}}$.

Let $\wh{z}$ denote a solution to variables $z$ in this linear program. The stochastic-update
strategy is defined to have 2 memory states (``transient" and ``recurrent"), with the transition
function defined to be
$\sigma_{\mathit{t}}(s)(a) = \wh{y}_a/\sum_{b\in \act{s}} \wh{y}_{b}$
and
$\sigma_{\mathit{r}}(s)(a) = \ov{x}_a/\sum_{b\in \act{s}} \ov{x}_{b}$,
and the probability of switching from ``transient" to ``recurrent" state upon entering $s$ being 
$\wh{y}_s/(\sum_{a\in \act{s}} \wh{y}_{a} + \wh{y}_s)$.
The correctness of the witness construction follows from~\cite{BBC+11}.

\smallskip\noindent{\bf MILP for memoryless strategies.}
For memoryless strategies, the current upper bound is NP~\cite{Cha07} and the previous algorithm enumerates all possible 
BSCCs under a memoryless strategy.
%%MECs\vojta{We enumerate MECS, too. Was BSCCs intended?}.
We present a polynomial-time reduction to solving a boolean combination of linear constraints,
that can be easily encoded using MILP with binary variables~\cite{Schrijver}. 
The key requirement for memoryless strategies is that a state can either be recurrent 
or transient.
For the existence question restricted to memoryless strategies we modify the linear 
constraints from Fig.~\ref{system-L} as follows: 
(i)~we add constraints; for all states $s$ and actions $b\in \act{s}$: 
$y_b>0 \implies (x_b>0 \lor \sum_{a \in \act{\src(b)}} x_a = 0)$; 
(ii)~we replace constraint (\ref{eq:yC}) from Fig.~\ref{system-L} by constraints 
that for all states $s$: $y_s = \sum_{a \in \act{s}} x_a$.
The constraint (ii) is a strengthening of constraint (\ref{eq:yC}), as 
the above constraint implies constraint (\ref{eq:yC}).
\ifthenelse{\isundefined{\techreport}}{}{
The intuition behind the additional constraint is as follows: 
$\sum_{a \in \act{\src(b)}} x_a = 0$ represents transient states where there is 
no restriction on $y_b$, otherwise $y_b$ can be positive only if $x_b$ is 
positive.
The witness strategy $\sigma$ is as follows:
let $\ov{z}$ denote a solution to the MILP for the memoryless strategy question;  
for a state $s$, if $\sum_{a \in \act{s}} \ov{x}_a=0$ 
(i.e., $s$ is transient), then $\sigma(s)(a)=  \ov{y}_a/\sum_{b\in \act{s}} \ov{y}_{b}$; 
otherwise, $\sigma(s)(a)=  \ov{x}_a/\sum_{b\in \act{s}} \ov{x}_{b}$.}
Further details are in~\ifthenelse{\isundefined{\techreport}}{\cite{techreport}}{Appendix}.

%\vojta{Explain how to get strategy for memoryless: we play according to $y_a$ until hitting a state
%where $\sum x_a$ is nonzero, and from this point on we play according to $x_a$.}

%TODO: In particular, we also check if an MDP is strongly connected, and then compute a memoryless witness strategy.\vojta{I need to implement this}

\smallskip\noindent{\bf Approximate Pareto curve for two objectives.}
To generate a Pareto curve, we successively compute solutions to several
linear programs for a single mean-payoff objective, where every time the objective is obtained
as a weighted sum of the objectives for which the Pareto curve is generated.
The weights are selected in a way similar to~\cite{FKP12}, 
%the one described in \cite{FKP12}, 
allowing us to obtain the approximation of the curve.

Unlike the PRISM implementation for multi-objective cumulative rewards, our tool is able to generate the Pareto curve
for objectives of the form \texttt{multi(R\{'r1'\}max=?[S], R\{'r2'\}max=? [S], R\{'r3'\}>=0.5 [S])} where the 
objectives to be optimised are subject to restrictions given by other rewards.

\smallskip\noindent{\bf Features of our tool.} 
In summary, our tool extends PRISM by developing algorithms to solve MDPs with 
multiple mean-payoff objectives.
Along with the algorithm from~\cite{BBC+11} we have also implemented
a visual representation of the Pareto curve for two-dimensional objectives. The implementation utilises
a multi-objective visualisation available in PRISM for cumulative reward and LTL objectives.

In addition, we adapted a feature from PRISM-games~\cite{CFK+13} which allows the user to generate strategies,
so that they can be explored and investigated by simulation. A product (Markov chain) of an MDP and a strategy can be
constructed, allowing the user to employ it for verification of other properties.

\begin{figure}[t]
\vspace{-1em}
\begin{center}
 \includegraphics[width=10cm]{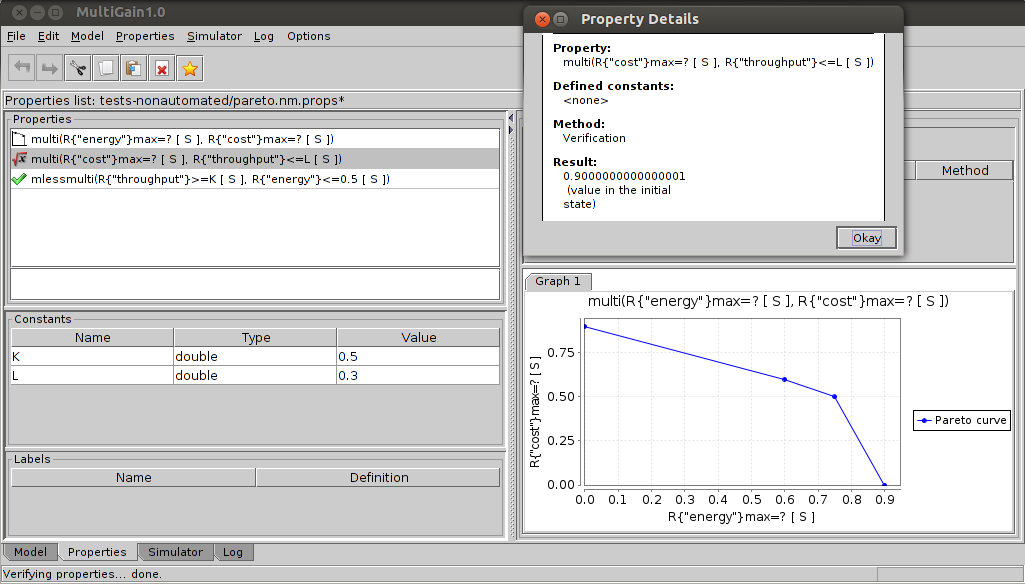}
\end{center}
 \caption{Screenshot of MultiGain (largely inheriting from the PRISM GUI).}
\vspace{-1em}
\end{figure}

 The tool is available
at~\url{http://qav.cs.ox.ac.uk/multigain/}, and the source code
is provided under GPL. For licencing reasons, Gurobi is not included with the download, but
it can be added manually by following provided steps.

\vspace{-1em}
\section{Experimental Results: Case Studies}
\vspace{-0.5em}
We have evaluated our tool on two standard case studies, adapted from \cite{prism-web}, and also
mention other applications where our tool could be used.

\smallskip
\noindent
{\em Dining philosophers} is a case study based on the algorithm of~\cite{DFP04},
which extends Lehmann and Rabin's randomised solution \cite{LR81} to the dining philosophers
problem so that there is no requirement for fairness assumptions. The constant $N$ gives the number of philosophers.
We use two reward structures, \texttt{think} and \texttt{eat} for the number of philosophers currently thinking and eating, respectively.

\smallskip
\noindent
{\em Randomised Mutual Exclusion} models a solution to the mutual exclusion problem by~\cite{Rab82}. The parameter $N$ gives the number of
processes competing for the access to the critical section.
Here we defined reward structures \texttt{try} and \texttt{crit} for the number of processes that are currently trying to access the critical
section, and those which are in it, currently (the latter number obviously never being more than 1).

\smallskip
\noindent
{\bf Evaluation}
The statistics for some of our experiments 
are given in Table~\ref{table:cs} (the complete results are available from the tool's website).
The experiments were run on a 2.66GHz PC with 4GB RAM, the LP solver used was Gurobi and the timeout (``t/o'') was set to 2 hours.
We observed that our approach scales to mid-size models, the main limitation being the LP solver.
%To be able to scale to larger models,
%we plan to integrate a different solution method, e.g. utilising the approach of \cite{symblicit} with \cite{FKP12}. 

{
%%\vspace{-3mm}
\begin{table}
\begin{center}
\scriptsize
\begin{tabular}{|c|c|c||c|c|c|c|c|c|}
\hline
\multirow{2}{*}{model}         & \multirow{2}{*}{para.} & property                      & MDP & \multicolumn{2}{c|}{LP} & total & solving & \multirow{2}{*}{value}\\
\cline{5-6}
         & & (A: \texttt{multi(}\dots\texttt{)}, B: \texttt{mlessmulti(}\dots\texttt{)})                      & states & vars\,(binary)& rows& time\,(s) & time\,(s) &\\
\hline
\hline
\multirow{4}{*}{phil}
              & $3$     & A:\,\texttt{R\{"think"\}max=?,\,R\{"eat"\}>=0.3} & 956       & 6344       & 1915      & 0.23      & 0.08       & 2.119 \\
              & $3$     & B:\,\texttt{R\{"think"\}>=2.11,\,R\{"eat"\}>=0.3} & 956       & 12553\,(6344)       & 11773      & 209.9      & 209.7       & true \\
              & $3$     & B:\,\texttt{R\{"think"\}>=2.12,\,R\{"eat"\}>=0.3} & 956       & 12553\,(6344)      & 11773      & 20.9      & 20.7       & false \\
              & $4$     & A:\,\texttt{R\{"think"\}max=?,\,R\{"eat"\}>=1}   & 9440      & 80368      & 18883     & 4.4      & 3.8       & 2.429 \\
              & $5$     & A:\,\texttt{R\{"think"\}max=?,\,R\{"eat"\}>=1}   & 93068     & 967168     & 186139    & 616.0     & 606.4     & 3.429\\
\hline
\hline
\multirow{3}{*}{mutex}
              & 3     & A:\,\texttt{R\{"try"\}max=?\,[S],\,R\{"crit"\}>=0.2}  & 27766     & 119038     & 55535     & 214.9     & 212.7     & 2.679\\
              & 4     & A:\,\texttt{R\{"try"\}max=?\,[S],\,R\{"crit"\}>=0.3}  & 668836    & 3010308    & 1337675   & {\em t/o}&{\em t/o}&{\em t/o}\\
              & 4     & A:\,\texttt{R\{"try"\}>=3.5\,[S],\,R\{"crit"\}>=0.3}    & 668836    & 3010308    & 1337676   & 4126       & 4073      & true\\
\hline
\end{tabular}
\end{center}
\caption{Experimental results. For space reasons, the \texttt{[S]} argument to \texttt{R} is omitted.\label{table:cs}}
\vspace{-8mm}
\end{table}}

\smallskip\noindent{\bf Other applications.}
We mention two applications which are solved using MDPs with multiple 
mean-payoff objectives.
(A)~The problem of synthesis from incompatible specifications was considered 
in~\cite{CGHRT12}.
Given a set of specifications $\varphi_1,\varphi_2,\ldots,\varphi_k$ 
that cannot be all satisfied together, the goal is to synthesize a system 
such that for all $1\leq i \leq k$ the distance to specification $\varphi_i$ is at most $v_i$.
In adversarial environments the problem reduces to games and for probabilistic environments to MDPs, 
with multiple mean-payoff objectives~\cite{CGHRT12}.
(B)~The problem of synthesis of steady state distributions for ergodic MDPs was
considered in~\cite{ABHH13}.
The problem can be modeled with multiple mean-payoff objectives by considering 
indicator reward functions $r_s$, for each state $s$, that assign reward $1$ to 
every action enabled in $s$ and~0 to all other actions.
The steady state distribution synthesis question of~\cite{ABHH13} then reduces to 
the existence question for multiple mean-payoff MDPs.
%MDPs with multiple mean-payoff objectives.

\smallskip\noindent{\bf Concluding remarks.} 
We presented the first tool for analysis of MDPs with multiple
mean-payoff objectives. 
The limiting factor is the LP solver,
%For MDPs with a single mean-payoff objective, a combination of explicit and symbolic 
%model-checking approaches has been developed in~\cite{symblicit}. 
and so an interesting direction would be to extend the results of~\cite{symblicit}
to multiple objectives.

%\clearpage
%
%\section{Comments from Vojtech}
%
%Technical details
%\begin{itemize}
% \item Extends PRISM \cite{KNP11}, in addition implementing the strategy generation feature of PRISM-games \cite{CFK+13}.
% \item Implemented using linear programming, LP solver we use is LPsolve \cite{lpsolve}
%\end{itemize}
%
%Features:
%\begin{itemize}
% \item Implementing algorithms from \cite{BBC+11} (for expected rewards only)
% \item Can solve arbitrary conjunction of mean payoff objectives
% \item Ability to draw a Pareto curve for 2 objectives
% \item Can get a strategy achieving a given formula (not for Pareto curve, though). The strategy can be saved, or used for simulation.
%\end{itemize}
%
%Some notes we might want to consider
%\begin{itemize}
% \item In the LICS paper we don't exactly say how the strategy can be generated, maybe it can be worth saying it here. What we do is we solve the program to get the answer whether the strategy exists, this gives us values for the $x_s$ variables. Then we construct a new program, where these values to $x_s$ variables give values for reaching $s$
% \item We should mention \cite{symblicit} which does single-objective mean-payoff. Not sure if it's in PRISM, the text does not say, but it's quite likely.
%\end{itemize}
%

\smallskip
\noindent
{\bf Acknowledgements} The authors were in part supported by
Austrian Science Fund (FWF) Grant No P23499- N23, FWF NFN Grant No S11407-N23 (RiSE), ERC Start grant (279307: Graph Games)
and the research centre Institute for Theoretical
Computer Science (ITI), grant No. P202/12/G061.

\vspace{-1em}
{\scriptsize
\bibliographystyle{abbrv}
\bibliography{main}
}
\ifthenelse{\isundefined{\techreport}}{}{
\clearpage

\section{Appendix}

\smallskip\noindent{\bf Argument for mixed ILP for memoryless strategies.}
The main idea of the correctness argument of the Boolean combination of 
linear constraints is as follows.
Given a memoryless strategy $\sigma$, once the strategy is fixed we obtain 
a Markov chain with two types of states, transient states and recurrent 
states. 
For recurrent states the variable $x$ represent the flow equations and 
intuitively, the variable $x_a$ denotes the long-run average frequency
of the action $a$.
The constraint $\sum_{a \in \act{s}} x_a=0$ is the constraint to represent
that a state is transient. 
The variables $y_a$ are used to determine the probabilities to reach the 
recurrent classes.
For an action $b$, if $\src(b)$ is transient, then for the first 
additional constraint for memoryless strategies, since 
$\sum_{a \in \act{\src(b)}} x_a=0$ is satisfied, it follows 
that there is no restriction on $y_b$.
Once a recurrent class is reached, it suffices to switch 
deterministically to the strategy of the recurrent class.
However, this is encoded slightly differently: we allow to 
continue before switching but only in the recurrent class, 
which is enforced by the constraint $y_b >0 \implies x_b >0$.
For a state $s$ in a recurrent class $\sum_{a \in \act{s}} x_a$
denotes the long-run average frequency of $s$, and the 
constraint (ii) requires that the long-run average frequency
of $s$ coincides with the probability $y_s$ to reach $s$ as 
the first state of the recurrent class.
Note that for a recurrent class $Z$, the sum $\sum_{s \in Z} y_s$ represents 
the probability that the recurrent class $Z$ is reached.
We describe the details of witness strategy constructions.

%\krish{Could you mention the following: Given a memoryless witness 
%strategy, what is the assignment to variables.
%Conversely, given assignment to variables, what is a witness memoryles strategy. 
%I started something below: not completely sure it matches your intuition.}

\smallskip\noindent{\em Strategy from solution.}
Given a solution to the mixed ILP, let $\ov{z}$ denote the solution to 
variable $z$. We construct a witness memoryless
strategy as follows: 
for a state $s$, if $\sum_{a \in \act{s}} \ov{x}_a > 0$, then for all $a'\in \act{s}$ 
the memoryless strategy plays $a'$ with probability $\ov{x}_{a'}/ \sum_{a \in \act{s}} \ov{x}_a$.
For a state $s$, if $\sum_{a \in \act{s}} \ov{x}_a = 0$, then for all $a'\in \act{s}$ 
the memoryless strategy plays $a'$ with probability $\ov{y}_{a'}/ \sum_{a \in \act{s}} \ov{y}_a$.

\smallskip\noindent{\em Solution from strategy.}
Consider a witness memoryless strategy $\sigma$ and consider the Markov chain 
obtained by fixing the strategy.
Let $X$ denote the set of recurrent states in the Markov chain, and 
$Y=S\setminus X$ the set of transient states.
The assignment of the variables to satisfy the constraints of the mixed ILP 
are as follows: 
(a)~For states $s$ in $X$ and action $a \in \act{s}$, the variable $x_{a}$ is 
assigned the long-run average frequency of action $a$ in the 
Markov chain; and $y_s$ is assigned the probability that the first state 
reached in the recurrent class containing $s$ is $s$.
(b)~For states $s$ in $Y$ and $a \in \act{s}$ we assign $x_a=0$ and $y_a=\sigma(s)(a)$ 
the probability assigned by the strategy $\sigma$.

}
\end{document}